\documentclass{article}

    \PassOptionsToPackage{numbers, compress}{natbib}

 \usepackage[main, final]{neurips_2026}

\usepackage[utf8]{inputenc} 
\usepackage[T1]{fontenc}    
\usepackage{hyperref}       
\usepackage{url}            
\usepackage{booktabs}       
\usepackage{amsfonts}       
\usepackage{nicefrac}       
\usepackage{microtype}      
\usepackage{xcolor}         
\usepackage{graphicx}       
\usepackage[most]{tcolorbox}

\title{GlyphAnchor: Enhancing Visual Text Rendering via Position-Anchored Glyph Priors}

\author{%
  \textbf{Qiang Xiang$^{1,2}$,\ Shuang Sun$^{2}$,\ Binglei Li$^{1,3}$,}\\
  \textbf{Yibo Chen$^{2}$, \ Xu Tang$^{2}$, Yao Hu$^{2}$, \ Junping Zhang$^{1}$\thanks{Corresponding author.}} \\
  $^1$Fudan University ~~\quad\quad $^2$Xiaohongshu Inc. ~~\quad\quad $^3$Shanghai Innovation Institute \\
  \texttt{\small \{qxiang24, blli24\}@m.fudan.edu.cn, jpzhang@fudan.edu.cn,}\\
  \texttt{\small \{sunshuang1, zhaohaibo, tangshen, xiahou\}@xiaohongshu.com}  \\
}

\begin{document}

\maketitle

\begin{figure}[ht]
\vspace{-25pt}
\centering
{\includegraphics[clip, trim=0cm 70cm 15cm 0cm, width=0.98\textwidth]{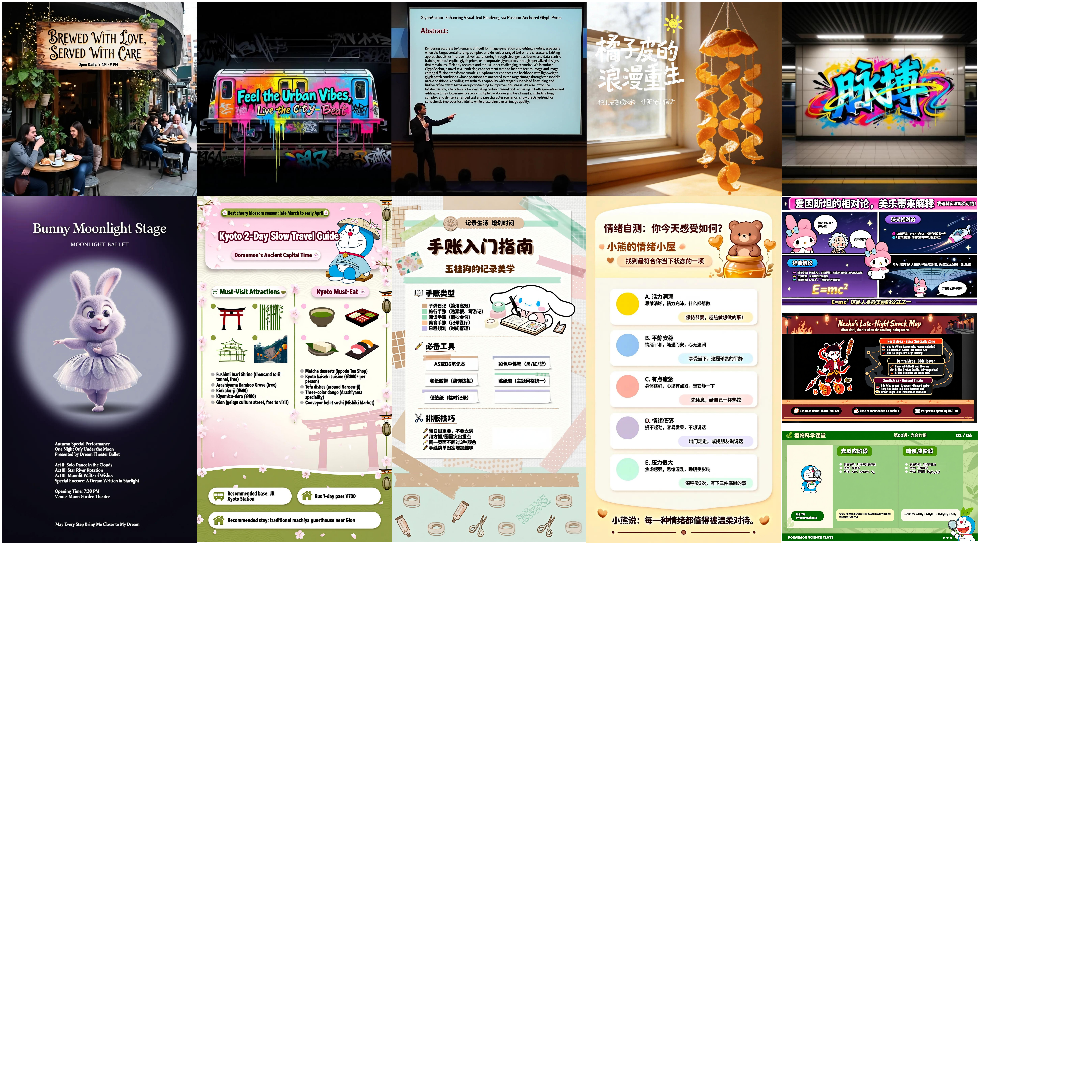}}
\caption{
\textbf{Generation results of GlyphAnchor.}}
\label{fig:teaser}
\end{figure}

\begin{abstract}
Rendering accurate text remains difficult for image generation and editing models, especially when the target contains long, complex, and densely arranged text or rare characters.
Existing approaches either improve native text rendering through stronger backbones and data-centric training without explicit glyph priors, or incorporate glyph priors through specialized designs that remain insufficiently accurate and robust under challenging scenarios.
We introduce GlyphAnchor, a novel text-rendering enhancement method for both text-to-image and image-editing diffusion transformer models.
GlyphAnchor enhances the backbone with lightweight glyph patch conditions whose positions are anchored to the target image through the model's native positional encoding.
We train this capability with staged supervised finetuning and further refine it with text-aware post-training to improve robustness.
We also introduce InfoTextBench, a benchmark for evaluating text-rich visual text rendering in both generation and editing settings.
Experiments across multiple backbones and benchmarks, including long, complex, and densely arranged text and rare character scenarios, show that GlyphAnchor consistently improves text fidelity while preserving overall image quality.
Code and models are publicly available at https://github.com/FireRedTeam/GlyphAnchor.
\end{abstract}

\section{Introduction}
Accurate visual text rendering has become a key bottleneck for modern text-to-image and image-editing foundation models. 
Recent image diffusion models~\citep{wu2025qwenimage,zhao2025lexart,zimageteam2025zimage,meituan2025longcatimage,baidu2026ernieimage} can synthesize high-quality scenes and follow complex instructions, but their text rendering remains less reliable for long, complex, and densely arranged text and rare characters.
These errors are especially damaging in posters, notes, infographics and other text-rich applications, where a single missing or wrong character can compromise usability.
This challenge arises in both text-to-image and image editing, and it involves three coupled requirements: the model must render the correct character shapes, place each text item at its intended location, and preserve the surrounding style and image quality.

Previous work has made substantial progress along several directions.
Glyph-prior methods incorporate rendered glyph images along with masks, glyph-aware encoders, or style controls into generation and editing models~\citep{ma2023glyphdraw,yang2023glyphcontrol,tuo2023anytext,tuo2024anytext2,liu2024glyphbyt5v2,zeng2024textctrl,wang2025glyphmastero,yan2025textmaster,jiang2025controltext,zhang2026freetext,yan2026glyphbanana}.
These methods show that explicit glyph information is useful for improving text accuracy.
Foundation models and data-centric methods~\citep{liu2023characteraware,wu2025qwenimage,zhao2025lexart,zimageteam2025zimage,meituan2025longcatimage,baidu2026ernieimage} improve native text rendering through stronger backbones, large-scale curated data, or character-aware training.
Other methods~\citep{chen2023textdiffuser,chen2023textdiffuser2,song2025dctext,shimoda2025typer,zhu2026textpecker,shuai2026glyphprinter} use layout-guided control, post-hoc correction, attention control, or reward-based optimization to improve text accuracy and repair generated text.
Despite these advances, accurate visual text rendering remains difficult for long, complex, and densely arranged text and rare characters.

To address this, we introduce \textbf{GlyphAnchor}, a method that improves visual text rendering with glyph priors.
GlyphAnchor organizes glyph priors as compact glyph patch images and anchors each patch to its intended location in the target image through RoPE positions~\citep{su2024roformer}.
Unlike methods that rely on custom attention masks, this design leaves the attention pattern unchanged and remains compatible with standard attention acceleration kernels such as FlashAttention~\citep{dao2022flashattention}.
It provides the backbone with explicit character-shape information and spatial cues, while keeping the conditioning mechanism lightweight and easy to adapt to different diffusion-transformer backbones.
We train this representation with staged supervised finetuning, which gradually transitions from ground-truth cropped patches to rendered glyph patches that match inference-time inputs.
We then apply text-aware DiffusionNFT post-training~\citep{zheng2025diffusionnft} to improve stability and visual quality in challenging cases.

Experiments show that GlyphAnchor consistently improves the corresponding backbones in challenging text-rendering scenarios, including long, complex, densely arranged text and rare characters.
To evaluate text-rich scenarios more directly, we build \textbf{InfoTextBench}, which measures word-level accuracy, phrase hit rate, aesthetics, and prompt following for both image editing and text-to-image settings.
On LongTextBench~\citep{geng2025xomni}, OneIGBench-text~\citep{chang2025oneig}, CVTG-2K~\citep{du2025textcrafter}, ChineseWord~\citep{wu2025qwenimage}, and InfoTextBench, GlyphAnchor consistently improves text accuracy, phrase hit rate, and overall visual quality relative to the corresponding backbones.
Our main contributions are listed below.

\qquad  1. We leverage glyph priors through a position-anchored design that provides glyph guidance with few extra tokens and is easy to adapt across diffusion-transformer backbones.

\qquad  2. We design a dedicated training recipe for position-anchored glyph priors, including staged supervised finetuning, text-aware rewards for NFT post-training, and train-time glyph augmentation.

\qquad  3. We build InfoTextBench for text-rich visual text rendering evaluation and show consistent improvements over the corresponding base models on benchmarks covering long, complex, and densely arranged text and rare characters.

\section{Related Work}

\textbf{Generation and editing with glyph priors.}
Many works improve visual text rendering by incorporating explicit glyph priors, such as rendered glyph images or glyph-aware text encoders, into the generation backbones. Some methods~\citep{ma2023glyphdraw,yang2023glyphcontrol,tuo2023anytext,tuo2024anytext2,ma2024glyphdraw2,Lu2025easytext,xie2026textflux,shi2025fonts} inject rendered glyph images, masks, positions, or text attributes as auxiliary conditions for generation.
Another line~\citep{zhao2024udifftext,liu2024glyphbyt5,liu2024glyphbyt5v2} improves text rendering by designing character-aware or glyph-aligned text encoders; and Glyph-ByT5-v2 is later adopted by foundation models such as Seedream 2.0~\citep{gong2025seedream2} and GLM-Image~\citep{zai2026glmimage}.
More recent work explores inference-time glyph injection: FreeText~\citep{zhang2026freetext} injects spectral glyph priors at inference time, and GlyphBanana~\citep{yan2026glyphbanana} injects glyph templates into the latent space and attention maps through an agentic workflow. For text editing, prior methods~\citep{zeng2024textctrl,wang2025glyphmastero,yan2025textmaster,lan2025fluxtext,jiang2025controltext,fang2025recognition} improve controllability through structure-style disentanglement, dedicated glyph encoders, or segmentation-based font control.

\textbf{Foundation models and data-centric methods.}
Another line emphasizes stronger foundation backbones, large-scale curated data, or character-aware training, without relying on an explicit inference-time glyph condition.
\citet{liu2023characteraware} argues that character-level text representations are key to improving spelling accuracy.
Qwen-Image~\citep{wu2025qwenimage}, Z-Image~\citep{zimageteam2025zimage}, LongCat-Image~\citep{meituan2025longcatimage}, ERNIE-Image~\citep{baidu2026ernieimage}, and X-Omni~\citep{geng2025xomni} demonstrate that large-scale foundation model training can achieve strong native text rendering and editing without any dedicated glyph module. LeX-Art~\citep{zhao2025lexart} highlights the importance of data synthesis quality and prompt enhancement for these approaches. 

\textbf{Layout-guided control, post-hoc correction, and RL.}
Some methods improve text rendering through layout-guided control rather than explicit glyph priors. TextDiffuser~\citep{chen2023textdiffuser,chen2023textdiffuser2} first predicts text layouts, converts them into segmentation masks, and conditions the diffusion model on these masks for accurate text placement. DCText~\citep{song2025dctext} and TextCrafter~\citep{du2025textcrafter} instead localize text regions during inference and handle them with attention mask binding or separate denoising. 
Type-R~\citep{shimoda2025typer} focuses on post-hoc correction by automatically detecting and repairing typographic errors in generated images. 
Other methods~\citep{zhu2026textpecker,shuai2026glyphprinter} optimize rendering quality through structural anomaly quantification or region-level DPO with local glyph-correctness annotations.

GlyphAnchor is complementary to these methods, focusing on how to leverage glyph priors in modern diffusion-transformer backbones.

\section{Method}
\label{sec:method}
To improve text rendering in modern text-to-image and image-editing foundation models, we incorporate explicit glyph priors that provide character-shape guidance beyond prompt tokens. Our goal is to inject these priors into diffusion-transformer backbones in a lightweight and seamless manner while preserving generation quality.

\subsection{Problem Formulation}

We consider both text-to-image and image editing models.
The input contains a prompt $p$ and an optional set of input images $\mathcal{I}_{\mathrm{src}}=\{I_{\mathrm{src}}\}$.
The output image should faithfully follow the prompt and render a set of target texts $\mathcal{S}=\{s_i\}_{i=1}^{N}$ specified in the prompt.
The texts may be short labels, long sentences, dense information blocks, or mixed Chinese, English, and symbols.

\subsection{Design Rationale}

To incorporate glyph priors into the backbone, we consider several design choices.
We use FireRed-Image-Edit-1.1, an image-editing model derived from Qwen-Image, as the backbone in this section because it performs better than Qwen-Image-Edit-2511 on the ChineseWord benchmark (Table~\ref{tab:chinese_word}).

\begin{figure}[t]
  \centering
  \includegraphics[clip, trim=0cm 100cm 0cm 0cm, width=0.98\textwidth]{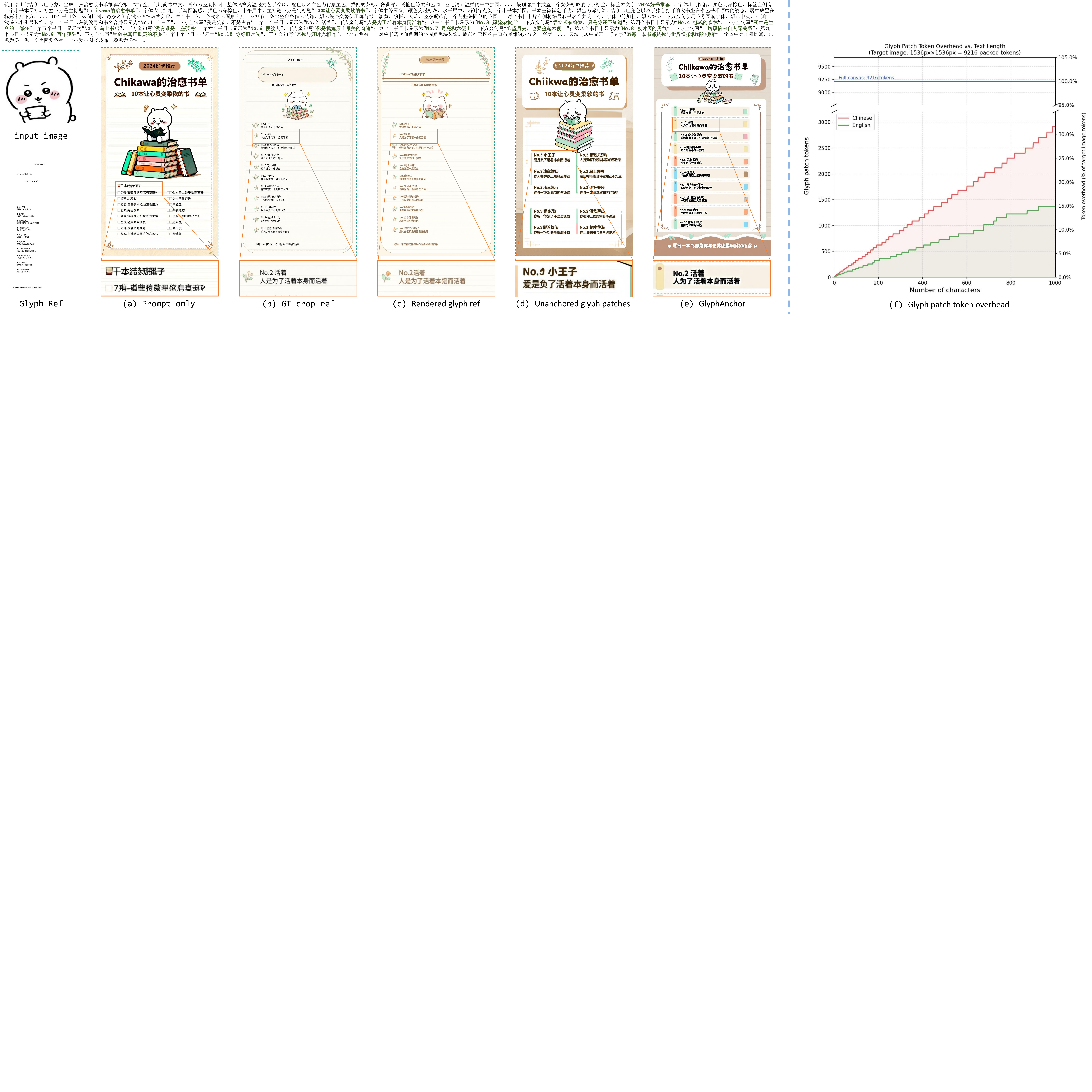}
  \caption{\textbf{Design choices for GlyphAnchor}, illustrated with FireRed-Image-Edit-1.1 examples. An English translation of the Chinese prompt is provided in Appendix~\ref{app:prompt_translation}.}
  \label{fig:glyph_prior_motivation}
\end{figure}

As illustrated in Fig.~\ref{fig:glyph_prior_motivation} (a), default prompt-only generation struggles to render long and complex text.
A straightforward design is to finetune the backbone with full-canvas glyph references, where all target texts are rendered on a blank canvas and fed as an additional input image (bottom left in Fig.~\ref{fig:glyph_prior_motivation}).
Fig.~\ref{fig:glyph_prior_motivation} (b) shows that this design can improve text accuracy, but it weakens prompt following and suppresses decorative details, background texture, and overall stylistic fidelity.
It is also expensive in token overhead, because a full-canvas glyph reference has the same spatial size as the target image and therefore contributes roughly one extra image frame of visual tokens.

To reduce this overhead and remove the unnecessary blank canvas, we next consider a more compact design that renders each target text as a glyph patch image and feeds these images as standard reference image inputs.
Fig.~\ref{fig:glyph_prior_motivation} (f) shows that even for long and complex text, glyph patches use far fewer tokens than full-canvas glyph references; for a 1536$\times$1536 target canvas, even 1000 Chinese or English characters occupy only about 31\% or 15\% of a full image frame, respectively.
However, compact glyph patches alone are not sufficient. Without explicit spatial grounding, the model can still miss text items or generate corrupted characters, as illustrated in Fig.~\ref{fig:glyph_prior_motivation} (d).
We therefore introduce a text layout representation:
\begin{equation}
\mathcal{L}=\{(s_i,b_i)\}_{i=1}^{N}, \qquad
b_i=(x_i,y_i,w_i,h_i),
\end{equation}
where $b_i$ is the normalized bounding box of $s_i$ on the target canvas.
The layout may be specified by the user or obtained from predefined templates, external layout tools, or an MLLM-based layout planner.
Our final design pairs compact glyph patch images with layout anchors, as shown in Fig.~\ref{fig:glyph_prior_motivation} (e). The full architecture is illustrated in Fig.~\ref{fig:glyphanchor_arch} and detailed in Sec.~\ref{sec:glyphanchor}.

Fig.~\ref{fig:glyph_prior_motivation} also reveals a train-test mismatch for glyph conditioning.
At inference, the target image is unknown, so glyph images must always be rendered from the target texts.
During training, by contrast, glyph images can instead be constructed from crops of the target image, which provide stronger supervision and lead to superior text accuracy.
However, this creates a train-test gap and biases the model toward copy-paste behaviour, degrading stylization, as reflected in Fig.~\ref{fig:glyph_prior_motivation} (b).
Training directly with rendered glyph images avoids this gap, but provides weaker supervision and leads to lower text accuracy, as illustrated in Fig.~\ref{fig:glyph_prior_motivation} (c).
These observations motivate our staged training strategy, as described in Sec.~\ref{sec:staged_sft}.

\subsection{GlyphAnchor}
\label{sec:glyphanchor}

Based on the above design rationale, \textbf{GlyphAnchor} adopts position-anchored glyph priors as follows.
For each text item $(s_i,b_i)$, we construct a glyph condition
\begin{equation}
c_i=(g_i, b_i), \qquad g_i = E(G_i),
\end{equation}
where $G_i$ is a glyph patch image and $E(\cdot)$ denotes VAE encoding into latent tokens.
During inference, $G_i$ is rendered from $s_i$ using the aspect ratio of $b_i$, with fixed line height and font size.

GlyphAnchor organizes all glyph patch images in a shared condition frame, which can be viewed as a virtual glyph plane aligned with the target latent grid.
Let the target latent grid have size $H\times W$.
For each text item, we map its normalized layout box $b_i=(x_i,y_i,w_i,h_i)$ to the $H\times W$ grid by scaling it to the target latent resolution and aligning the center of $g_i$ with the box center, thereby obtaining the position IDs of the glyph patch tokens.

\begin{figure}[t]
  \centering
  \includegraphics[clip, trim=5cm 73cm 2cm 0cm, width=0.95\textwidth]{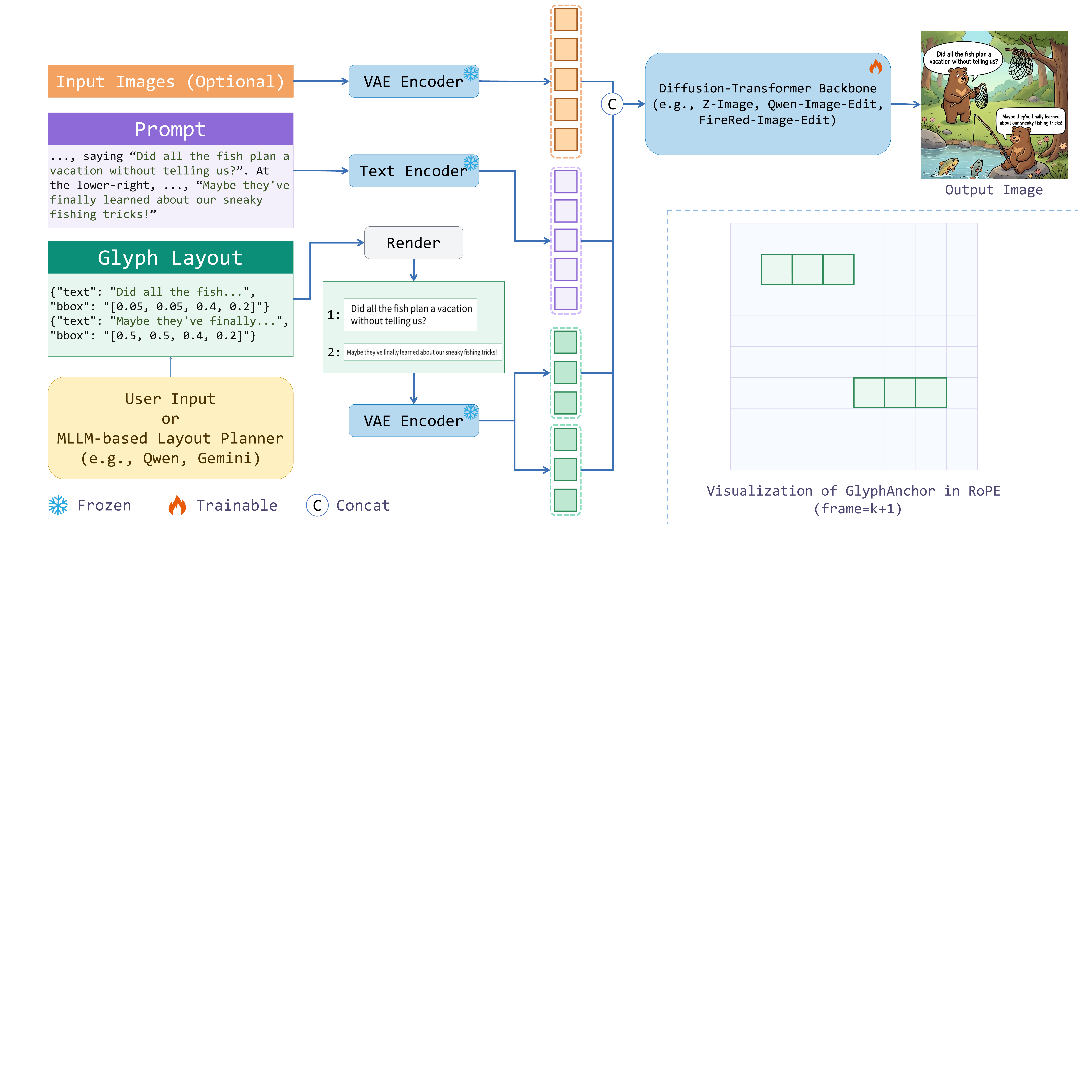}
  \caption{\textbf{Overview of GlyphAnchor}. Prompt, optional source images, and rendered glyph patch images are encoded into tokens and concatenated before a text-to-image or image-editing diffusion-transformer backbone. The resulting glyph patch tokens are placed in an additional condition frame, forming a virtual glyph plane whose coordinates are anchored to the target layout.}
  \label{fig:glyphanchor_arch}
\end{figure}

At inference, layouts are either specified by the user or planned by an MLLM, followed by post-processing such as deduplication, overlap adjustment, and text wrapping.
The glyph patch tokens are then concatenated with the standard visual token sequence, but their position IDs follow the layout anchors described above rather than the backbone's default positional assignment.
We place the glyph conditions in a virtual glyph plane after the noisy target frame and any optional source-image frames.
As shown in Fig.~\ref{fig:glyphanchor_arch}, these tokens are then processed jointly by the text-to-image or image-editing backbone.
Because many recent diffusion-transformer image generation models~\citep{wu2025qwenimage,zimageteam2025zimage,blackforestlabs2025flux2dev} use 3D RoPE or its variants, GlyphAnchor can be easily adapted to them by only adding glyph tokens and reassigning their position IDs.
For classifier-free guidance, the conditional branch receives the glyph conditions and the unconditional branch omits them.

\subsection{Staged Supervised Finetuning for GlyphAnchor}
\label{sec:staged_sft}

We use staged supervised finetuning to adapt the backbone to GlyphAnchor.
Let $z_t=(1-\sigma_t)z_0+\sigma_t\epsilon$ denote the noisy image latent at timestep $t$ under the flow-matching schedule, where $\epsilon\sim\mathcal{N}(0,I)$, and let $v_t=\epsilon-z_0$ be the velocity target.
Given the noisy target tokens, prompt embedding, optional source-image tokens, and position-anchored glyph patch tokens, the transformer is trained with
\begin{equation}
\mathcal{L}_{\mathrm{sft}}
= \mathbb{E}_{\epsilon,\, z_0,\, t,\, p,\, \mathcal{I}_{\mathrm{src}},\, \{c_i\}}
\left[
\left\|
\hat{v}_{\theta}(z_t, t, p, \mathcal{I}_{\mathrm{src}}, \{c_i\})
- v_t
\right\|_2^2
\right].
\label{eq:sft_loss}
\end{equation}

We use three types of glyph conditions during training: ground-truth image crops within $b_i$, rendered glyph patch images whose size follows the target box, and rendered glyph patch images with fixed line height and font size.
Their sampling ratios are scheduled across training stages.
Across stages, the sampling weight gradually shifts from ground-truth crops toward rendered glyph patches.
This progression bridges the train-test gap. Ground-truth crops provide stronger supervision but are unavailable at inference, while rendered glyph patch images match inference-time inputs but are harder to learn from directly.
During training, we further apply perturbation-based augmentation to the rendered glyph patch images, including random line breaks, resizing, rotation, translation, and dropout.
These augmentations prevent the model from overfitting to the exact rendered appearance of glyph patches, while preserving their character-shape and coarse layout cues.

\subsection{NFT Post-training with Text-Aware Rewards}
\label{sec:nft}
After supervised finetuning, the model can still suffer from limited stability and a relatively high failure rate on difficult cases.
We therefore adopt the DiffusionNFT framework~\cite{zheng2025diffusionnft} for post-training to improve stability.
We further design three text-aware rewards for text correctness, style consistency, and visual fusion.

\textbf{Text-Layout Consistency Reward.}
String-level OCR rewards, typically based on Levenshtein edit distance between recognized text and target text, can be exploited. A model may increase OCR accuracy by producing oversized characters that are easier to recognize. Also, the edit-distance objective does not penalize wrong layout in generated images.
We therefore score text correctness and spatial consistency jointly.
We parse the OCR outputs of the generated image and the ground-truth image into character instances with labels, locations, and scales.
The textual term $R_{\mathrm{text}}$ is computed as one minus the normalized Levenshtein edit distance between the recognized text $s_{\mathrm{pred}}$ and the target text $s_{\mathrm{tgt}}$.
We then form a set $\mathcal{M}$ of matched character pairs between characters from the generated image and those from the ground-truth image, where each pair shares the same character label, and compute a spatial term to penalize center displacement and scale mismatch:
\begin{equation}
R_{\mathrm{layout}}
=
\frac{\gamma_{\mathrm{cov}}}{|\mathcal{M}|}
\sum_{i\in\mathcal{M}}
\exp(-\lambda d_i-\alpha \Delta s_i),
\qquad
\gamma_{\mathrm{cov}}=\frac{|\mathcal{M}|}{|s_{\mathrm{tgt}}|}.
\end{equation}
Here $\gamma_{\mathrm{cov}}$ is a character-level coverage term, $d_i$ is the normalized distance between the $i$-th matched character pair, and $\Delta s_i$ measures the difference between their character scales.
Then, we define:
\begin{equation}
R_{\mathrm{TLC}}
=
w_{\mathrm{text}}\,R_{\mathrm{text}}
+
w_{\mathrm{layout}}\,\mathrm{Gate}(R_{\mathrm{text}})\,R_{\mathrm{layout}},
\end{equation}
where $\mathrm{Gate}(\cdot)$ is a thresholding function that activates the layout term only when text recognition is sufficiently accurate.

\textbf{Style and fusion rewards.}
Text correctness alone is insufficient; the rendered text should also match the intended typography and blend with the surroundings seamlessly.
We use $R_{\mathrm{style}}$ to measure whether the generated text style matches the target, and $R_{\mathrm{fusion}}$ to measure whether the text is naturally integrated with local background and visual context.
Both rewards are scored by an MLLM judge (Qwen3.5-122B-A10B~\citep{qwen3.5} in our implementation).
The final reward is
\begin{equation}
R =
\lambda_{\mathrm{TLC}} R_{\mathrm{TLC}}
+\lambda_{\mathrm{style}}R_{\mathrm{style}}
+\lambda_{\mathrm{fusion}}R_{\mathrm{fusion}}.
\end{equation}

\subsection{Training Data}

Our training data is composed of several curated datasets spanning both image editing and text-to-image.
It covers a broad range of scenarios, including web-curated text-rich images and template-constructed images. In total, the final training mixture contains approximately 500K samples.
We control the sampling ratios across these sources during training.
Overall, web-curated data and template-constructed data are sampled at an approximately balanced ratio, while the overall mixture of image editing and text-to-image data is maintained at roughly 3:1.

\subsection{InfoTextBench}
\label{sec:infotextbench}
Although existing benchmarks such as LongTextBench~\citep{geng2025xomni} evaluate long-text rendering, they do not adequately cover text-rich scenarios with densely arranged and highly structured content. We therefore build \textbf{InfoTextBench}.
InfoTextBench contains 133 samples, with Chinese and English prompt versions for each.
Each sample includes a web-collected reference image, manually curated paired image-editing and text-to-image prompts, and an explicit target text list.
Both prompts describe the same poster content, with the editing prompt relying on the reference image and the text-to-image prompt describing it explicitly.
The prompts are long and detailed, averaging 1065 and 1127 prompt words for image editing and text-to-image, respectively.
Each sample contains 19.3 target text items on average, with an average total text length of 151 words, often mixing Chinese, English, and symbols.
The content covers diverse text-rich visual layouts across multiple real-world scenarios.
Compared with LongTextBench, InfoTextBench contains substantially longer target texts, as quantified in Appendix~\ref{app:benchmark_comparison}.

\textbf{Evaluation protocol.}
We evaluate generated results along two dimensions: text accuracy and overall visual quality.
For text accuracy, we use PaddleOCR~\citep{cui2025paddleocr3} and compare the recognized text against the target text list.
We report normalized edit distance (NED, defined as $1$ minus the normalized Levenshtein edit distance so that higher is better), together with word-level recall, word-level precision, word-level F1, and phrase hit rate.
Here, word-level recall measures how much target text is correctly generated, while word-level precision measures how much generated text is correct rather than wrong or extra. We use word-level F1 to summarize them.
Phrase hit rate measures the fraction of target phrases found in the generated image, making it more sensitive to multi-block poster content.
For overall visual quality, we use an MLLM judge to score aesthetics and prompt following separately, and take their average as the final overall visual quality score.

\section{Experiments}
\label{sec:experiments}

We evaluate GlyphAnchor on both text-to-image and image-editing backbones across general text-rendering benchmarks, ChineseWord for rare characters, and our text-rich InfoTextBench.

\subsection{Experimental Setup}

\textbf{Implementation details.}
We implemented GlyphAnchor on three diffusion-transformer backbones: FireRed-Image-Edit-1.1~\citep{fireredteam2026fireredimageedit} and Qwen-Image-Edit-2511~\citep{wu2025qwenimage} for image editing, and Z-Image~\citep{zimageteam2025zimage} for text-to-image. 
For each backbone, we first conduct staged supervised finetuning for two days on 24 H800-80GB GPUs using AdamW~\citep{adamw} with a learning rate of $5\times10^{-6}$.
We then perform NFT post-training for another two days on the same hardware setup, using AdamW with a learning rate of $3\times10^{-4}$.
For NFT, we sample 8 images per prompt and set the number of groups to 12, with $(\lambda_{\mathrm{TLC}},\lambda_{\mathrm{style}},\lambda_{\mathrm{fusion}})=(0.5,0.25,0.25)$.

\textbf{Evaluation setup.}
The evaluation covers LongTextBench~\citep{geng2025xomni}, OneIGBench-text~\citep{chang2025oneig}, CVTG-2K~\citep{du2025textcrafter}, ChineseWord~\citep{wu2025qwenimage}, and our InfoTextBench, which together test long, complex, and densely arranged text, as well as rare characters.
Since GlyphAnchor requires a layout input, we use Qwen3.5-122B-A10B~\citep{qwen3.5} as the MLLM-based layout planner to generate bounding boxes for the target text items.
For ChineseWord, we use the \textit{Table of General Standard Chinese Characters}~\citep{statecouncil2013hanzi} as the source vocabulary, which includes 3,500 Level-1 commonly used characters, 3,000 Level-2 characters, and 1,605 Level-3 characters. We evaluate character rendering by generating one image for each target character. 
For InfoTextBench, we generate four images for each prompt and report the average score over all generated images.
All metrics are higher-is-better unless stated otherwise.

\definecolor{gainpurple}{RGB}{112,48,160}

\newcommand{\tabmaintextbench}{
\begin{table}[t]
  \centering
  \scriptsize
  \setlength{\tabcolsep}{2pt}
  \caption{\textbf{Quantitative comparison on LongTextBench, OneIGBench-text, and CVTG-2K.}}
  \label{tab:main_text_bench}
  \resizebox{0.85\textwidth}{!}{
  \begin{tabular}{lccccccccc}
    \toprule
    Method &
    \multicolumn{3}{c}{LongTextBench} &
    \multicolumn{3}{c}{OneIGBench-text} &
    \multicolumn{3}{c}{CVTG-2K} \\
    \cmidrule(lr){2-4}\cmidrule(lr){5-7}\cmidrule(lr){8-10}
    & EN & ZH & Avg. & EN & ZH & Avg. & NED & CLIP & WAC \\
    \midrule
    \multicolumn{10}{l}{\textit{Closed-source models}} \\
    GPT-Image-1~\citep{openai2025gptimage1} & 0.956 & 0.619 & 0.788 & 0.857 & 0.650 & 0.754 & 0.948 & 0.798 & 0.857 \\
    Seedream 3.0~\citep{gao2025seedream3} & 0.896 & 0.878 & 0.887 & 0.865 & 0.928 & 0.897 & 0.854 & 0.782 & 0.592 \\
    Seedream 4.0~\citep{seedreamteam2025seedream4} & 0.921 & 0.926 & 0.924 & 0.981 & 0.982 & 0.982 & 0.922 & 0.798 & 0.845 \\
    Nano Banana 2.0~\citep{google2026nanobanana2} & 0.981 & 0.949 & 0.965 & 0.944 & 0.983 & 0.964 & 0.875 & 0.737 & 0.779 \\
    \midrule
    \multicolumn{10}{l}{\textit{Open-source baselines}} \\
    TextCrafter~\citep{du2025textcrafter} & -- & -- & -- & -- & -- & -- & 0.868 & 0.787 & 0.737 \\
    TextDiffuser-2~\citep{chen2023textdiffuser2} & -- & -- & -- & -- & -- & -- & 0.435 & 0.677 & 0.233 \\
    AnyText~\citep{tuo2023anytext} & -- & -- & -- & -- & -- & -- & 0.468 & 0.743 & 0.180 \\
    Qwen-Image~\citep{wu2025qwenimage} & 0.943 & 0.946 & 0.945 & 0.891 & 0.963 & 0.927 & 0.912 & 0.802 & 0.829 \\
    Qwen-Image-2512~\citep{wu2025qwenimage} & 0.956 & 0.965 & 0.960 & \underline{0.988} & 0.984 & 0.986 & 0.929 & 0.782 & 0.860 \\
    FLUX.2-klein-9B~\citep{blackforestlabs2026flux2klein} & 0.864 & 0.218 & 0.541 & 0.866 & 0.492 & 0.679 & -- & -- & -- \\
    Emu3.5~\citep{cui2025emu35} & 0.976 & 0.928 & 0.952 & \textbf{0.994} & 0.941 & 0.968 & 0.966 & -- & 0.912 \\
    GLM-Image~\citep{zai2026glmimage} & 0.952 & 0.979 & 0.966 & 0.969 & 0.976 & 0.973 & 0.956 & 0.788 & 0.912 \\
    LongCat-Image~\citep{meituan2025longcatimage} & 0.871 & 0.953 & 0.912 & 0.965 & 0.945 & 0.955 & 0.936 & 0.786 & 0.866 \\
    ERNIE-Image~\citep{baidu2026ernieimage} & 0.968 & 0.959 & 0.964 & 0.967 & 0.898 & 0.932 & -- & -- & -- \\
    Z-Image-Turbo~\citep{zimageteam2025zimage} & 0.917 & 0.926 & 0.922 & \textbf{0.994} & 0.982 & \underline{0.988} & 0.928 & \underline{0.805} & 0.859 \\
    \midrule
    \multicolumn{10}{l}{\textit{GlyphAnchor final models}} \\
    FireRed-Image-Edit-1.1~\citep{fireredteam2026fireredimageedit} & 0.895 & 0.898 & 0.897 & 0.813 & 0.937 & 0.875 & 0.865 & 0.785 & 0.747 \\
    \quad + GlyphAnchor & \textbf{0.990} & 0.980 & \textbf{0.985} & 0.961 & 0.969 & 0.965 & \underline{0.968} & 0.782 & 0.917 \\
     & \textcolor{gainpurple}{\textit{(+.095)}} & \textcolor{gainpurple}{\textit{(+.082)}} & \textcolor{gainpurple}{\textit{(+.088)}} & \textcolor{gainpurple}{\textit{(+.148)}} & \textcolor{gainpurple}{\textit{(+.032)}} & \textcolor{gainpurple}{\textit{(+.090)}} & \textcolor{gainpurple}{\textit{(+.103)}} & \textcolor{gainpurple}{\textit{(-.003)}} & \textcolor{gainpurple}{\textit{(+.170)}} \\
    Qwen-Image-Edit-2511~\citep{wu2025qwenimage} & 0.900 & 0.924 & 0.912 & 0.893 & 0.963 & 0.928 & 0.850 & 0.774 & 0.717 \\
    \quad + GlyphAnchor & \underline{0.983} & \underline{0.984} & \underline{0.984} & 0.960 & 0.975 & 0.967 & \underline{0.968} & 0.801 & \underline{0.922} \\
     & \textcolor{gainpurple}{\textit{(+.083)}} & \textcolor{gainpurple}{\textit{(+.060)}} & \textcolor{gainpurple}{\textit{(+.072)}} & \textcolor{gainpurple}{\textit{(+.067)}} & \textcolor{gainpurple}{\textit{(+.012)}} & \textcolor{gainpurple}{\textit{(+.039)}} & \textcolor{gainpurple}{\textit{(+.118)}} & \textcolor{gainpurple}{\textit{(+.027)}} & \textcolor{gainpurple}{\textit{(+.205)}} \\
    Z-Image~\citep{zimageteam2025zimage} & 0.935 & 0.936 & 0.936 & 0.987 & \underline{0.988} & \underline{0.988} & 0.937 & 0.797 & 0.867 \\
    \quad + GlyphAnchor & 0.975 & \textbf{0.987} & 0.981 & \underline{0.988} & \textbf{0.994} & \textbf{0.991} & \textbf{0.970} & \textbf{0.823} & \textbf{0.924} \\
     & \textcolor{gainpurple}{\textit{(+.040)}} & \textcolor{gainpurple}{\textit{(+.051)}} & \textcolor{gainpurple}{\textit{(+.045)}} & \textcolor{gainpurple}{\textit{(+.001)}} & \textcolor{gainpurple}{\textit{(+.006)}} & \textcolor{gainpurple}{\textit{(+.003)}} & \textcolor{gainpurple}{\textit{(+.033)}} & \textcolor{gainpurple}{\textit{(+.026)}} & \textcolor{gainpurple}{\textit{(+.057)}} \\
    \bottomrule
  \end{tabular}
  }
\end{table}
}

\newcommand{\tabablation}{
\begin{table}[t]
  \centering
  \small
  \setlength{\tabcolsep}{3pt}
  \caption{\textbf{Ablation study on the FireRed-Image-Edit-1.1 backbone.}}
  \label{tab:ablation}
  \resizebox{\textwidth}{!}{
  \begin{tabular}{@{}lccccccccccc}
    \toprule
    Variant &
    \multicolumn{3}{c}{LongTextBench} &
    \multicolumn{6}{c}{InfoTextBench} \\
    \cmidrule(lr){2-4}\cmidrule(lr){5-10}
    & EN & ZH & Avg. & NED & Word Recall & Word Precision & Word F1 & Phrase Hit & Overall Visual Quality \\
    \midrule
    FireRed-Image-Edit-1.1 & 0.895 & 0.898 & 0.897 & 0.210 & 0.443 & 0.463 & 0.433 & 0.154 & 4.608 \\
    \quad + default SFT with train data & 0.907 & 0.905 & 0.906 & 0.242 & 0.501 & 0.537 & 0.510 & 0.284 & 4.931 \\
    \quad + glyph patch SFT w/o anchor & 0.925 & 0.935 & 0.930 & 0.416 & 0.794 & 0.701 & 0.737 & 0.399 & 5.012 \\
    \quad + glyph patch SFT w/ anchor & \underline{0.984} & \underline{0.974} & \underline{0.979} & \underline{0.647} & \underline{0.949} & \underline{0.856} & \underline{0.892} & \underline{0.711} & \underline{6.395} \\
    \quad + text-aware NFT & \textbf{0.990} & \textbf{0.980} & \textbf{0.985} & \textbf{0.651} & \textbf{0.950} & \textbf{0.864} & \textbf{0.898} & \textbf{0.713} & \textbf{6.480} \\
    \bottomrule
  \end{tabular}
  }
\end{table}
}

\newcommand{\tabinfotextbench}{
\begin{table}[t]
  \centering
  \small
  \setlength{\tabcolsep}{4pt}
  \caption{\textbf{Quantitative comparison on InfoTextBench.}}
  \label{tab:info_text_bench}
  \resizebox{\textwidth}{!}{
  \begin{tabular}{lcccccccc}
    \toprule
    Method & NED & Word Recall & Word Precision & Word F1 & Phrase Hit & Overall Visual Quality \\
    \midrule
    \multicolumn{7}{l}{\textit{Editing models}} \\
    Seedream 4.5~\citep{byteplus2025seedream45} & 0.432 & 0.865 & 0.699 & 0.765 & 0.674 & 6.360 \\
    Nano Banana Pro~\citep{google2025nanobananapro} & 0.649 & \textbf{0.978} & 0.916 & \underline{0.942} & \textbf{0.928} & \textbf{9.254} \\
    FLUX.2-dev~\citep{blackforestlabs2025flux2dev} & 0.226 & 0.505 & 0.529 & 0.505 & 0.116 & 4.311 \\
    GLM-Image~\citep{zai2026glmimage} & 0.505 & 0.719 & \textbf{0.938} & 0.803 & 0.583 & 5.787 \\
    LongCat-Image Edit~\citep{meituan2025longcatimage} & 0.253 & 0.445 & 0.704 & 0.505 & 0.278 & 4.477 \\
    \midrule
    FireRed-Image-Edit-1.1~\citep{fireredteam2026fireredimageedit} & 0.210 & 0.443 & 0.463 & 0.433 & 0.154 & 4.608 \\
    \quad + GlyphAnchor & \underline{0.651} & 0.950 & 0.864 & 0.898 & 0.713 & 6.480 \\
     & \textcolor{gainpurple}{\textit{(+.441)}} & \textcolor{gainpurple}{\textit{(+.507)}} & \textcolor{gainpurple}{\textit{(+.401)}} & \textcolor{gainpurple}{\textit{(+.465)}} & \textcolor{gainpurple}{\textit{(+.559)}} & \textcolor{gainpurple}{\textit{(+1.872)}} \\
    Qwen-Image-Edit-2511~\citep{wu2025qwenimage} & 0.237 & 0.466 & 0.543 & 0.479 & 0.167 & 4.603 \\
    \quad + GlyphAnchor & 0.632 & \underline{0.957} & 0.817 & 0.871 & 0.737 & 6.704 \\
     & \textcolor{gainpurple}{\textit{(+.395)}} & \textcolor{gainpurple}{\textit{(+.491)}} & \textcolor{gainpurple}{\textit{(+.274)}} & \textcolor{gainpurple}{\textit{(+.392)}} & \textcolor{gainpurple}{\textit{(+.570)}} & \textcolor{gainpurple}{\textit{(+2.101)}} \\
    \midrule
    \multicolumn{7}{l}{\textit{Text-to-image models}} \\
    Qwen-Image~\citep{wu2025qwenimage} & 0.234 & 0.421 & 0.760 & 0.490 & 0.248 & 4.910 \\
    Qwen-Image-2512~\citep{wu2025qwenimage} & 0.500 & 0.802 & 0.794 & 0.788 & 0.546 & 6.344 \\
    LongCat-Image~\citep{meituan2025longcatimage} & 0.242 & 0.432 & 0.740 & 0.494 & 0.295 & 4.474 \\
    \midrule
    Z-Image~\citep{zimageteam2025zimage} & 0.566 & 0.949 & 0.821 & 0.872 & 0.807 & 7.092 \\
    \quad + GlyphAnchor & \textbf{0.710} & \textbf{0.978} & \underline{0.929} & \textbf{0.950} & \underline{0.885} & \underline{8.822} \\
     & \textcolor{gainpurple}{\textit{(+.144)}} & \textcolor{gainpurple}{\textit{(+.029)}} & \textcolor{gainpurple}{\textit{(+.108)}} & \textcolor{gainpurple}{\textit{(+.078)}} & \textcolor{gainpurple}{\textit{(+.078)}} & \textcolor{gainpurple}{\textit{(+1.730)}} \\
    \bottomrule
  \end{tabular}
  }
\end{table}
}

\newcommand{\tabchineseword}{
\begin{table}[t]
  \centering
  \small
  \setlength{\tabcolsep}{8pt}
  \caption{\textbf{Quantitative comparison on ChineseWord.}}
  \label{tab:chinese_word}
  \resizebox{0.55\textwidth}{!}{
  \begin{tabular}{lcccc}
    \toprule
    Method & Level-1 & Level-2 & Level-3 & Total \\
    \midrule
    Qwen-Image~\citep{wu2025qwenimage} & 0.937 & 0.332 & 0.026 & 0.533 \\
    Qwen-Image-2512~\citep{wu2025qwenimage} & 0.936 & 0.382 & 0.039 & 0.553 \\
    LongCat-Image~\citep{meituan2025longcatimage} & 0.907 & 0.714 & 0.379 & 0.731 \\
    \midrule
    FireRed-Image-Edit-1.1~\citep{fireredteam2026fireredimageedit} & 0.926 & 0.411 & 0.067 & 0.565 \\
    \quad + GlyphAnchor & \textbf{0.992} & \underline{0.783} & \underline{0.416} & \underline{0.801} \\
     & \textcolor{gainpurple}{\textit{(+.066)}} & \textcolor{gainpurple}{\textit{(+.372)}} & \textcolor{gainpurple}{\textit{(+.349)}} & \textcolor{gainpurple}{\textit{(+.236)}} \\
    Qwen-Image-Edit-2511~\citep{wu2025qwenimage} & 0.816 & 0.360 & 0.051 & 0.496 \\
    \quad + GlyphAnchor & \underline{0.987} & \textbf{0.795} & \textbf{0.438} & \textbf{0.807} \\
     & \textcolor{gainpurple}{\textit{(+.171)}} & \textcolor{gainpurple}{\textit{(+.435)}} & \textcolor{gainpurple}{\textit{(+.387)}} & \textcolor{gainpurple}{\textit{(+.311)}} \\
    Z-Image~\citep{zimageteam2025zimage} & 0.852 & 0.278 & 0.025 & 0.476 \\
    \quad + GlyphAnchor & 0.939 & 0.377 & 0.102 & 0.565 \\
     & \textcolor{gainpurple}{\textit{(+.087)}} & \textcolor{gainpurple}{\textit{(+.099)}} & \textcolor{gainpurple}{\textit{(+.077)}} & \textcolor{gainpurple}{\textit{(+.089)}} \\
    \bottomrule
  \end{tabular}
  }
\end{table}
}

\tabmaintextbench
\tabchineseword
\tabinfotextbench

\subsection{Quantitative Results}

\textbf{General text-rendering benchmarks.}
Table~\ref{tab:main_text_bench} shows that GlyphAnchor consistently improves long text rendering across all three backbones and three evaluated benchmarks.
The gains are particularly pronounced for the two editing models, while the already strong Z-Image baseline still benefits on both LongTextBench and CVTG-2K.
Notably, Z-Image-based GlyphAnchor achieves the best CVTG-2K results among all compared methods, indicating that explicit glyph anchoring improves text fidelity while maintaining semantic alignment.

\textbf{Rare character rendering.}
Table~\ref{tab:chinese_word} further shows that GlyphAnchor is especially effective on rare character rendering, where prompts alone are insufficient.
Across all backbones, GlyphAnchor improves rendering capability for both common (Level 1) and rare (Level 2 and 3) characters.
This pattern supports our central hypothesis: explicit glyph evidence becomes most valuable when the prompt alone cannot reliably specify the glyph structure of the target characters.

\textbf{Text-rich generation and editing.}
Table~\ref{tab:info_text_bench} evaluates models on InfoTextBench, a more demanding benchmark where models must render many text items while preserving layout and visual coherence.
GlyphAnchor improves both word-level and phrase-level text accuracy, as well as overall visual quality, for both editing and text-to-image models.
These results suggest that glyph anchoring helps not only local text correctness but also the practical usability of text-rich images.

\subsection{Qualitative Results}

\begin{figure}[t]
  \centering
  {\includegraphics[clip, trim=0cm 18cm 19.5cm 0cm, width=0.98\textwidth]{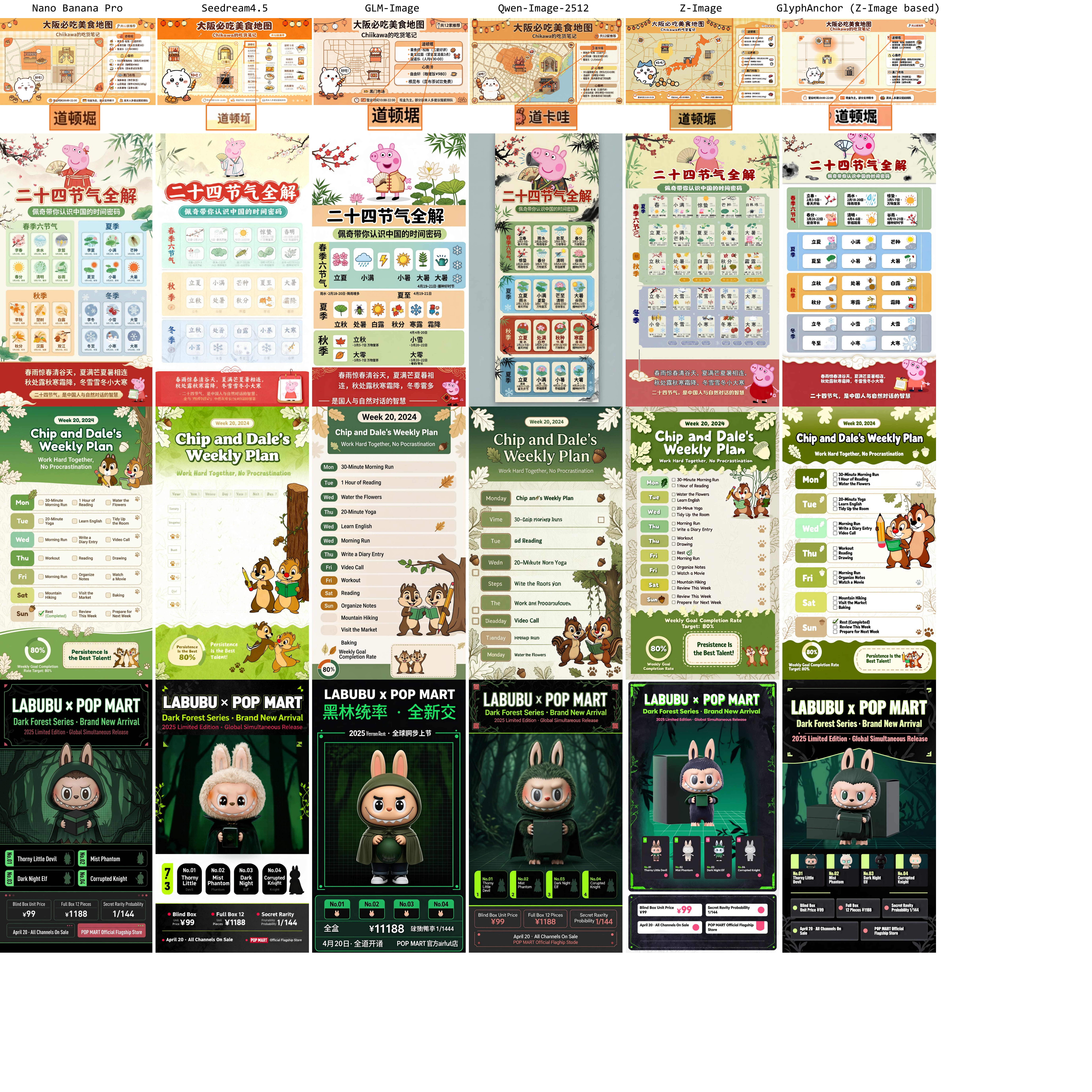}}
  \caption{\textbf{Qualitative comparison on challenging text-rich cases.} Best viewed with zoom.}
  \label{fig:qualitative}
  \vspace{-3pt}
\end{figure}

Fig.~\ref{fig:qualitative} presents qualitative comparisons on text-rich cases, where models must render long titles, dense small text, and structured layouts.
While closed-source Nano Banana Pro and GlyphAnchor both render these examples well, other baselines still suffer from common text-rendering errors, such as corrupted characters, missing or duplicated words, illegible small text, and incorrect layout.
For example, in the third row, Seedream 4.5 fails to follow the prompt-specified layout, GLM-Image also produces an incorrect layout, Qwen-Image-2512 contains missing, incorrect, and corrupted words, and Z-Image omits some words.
These comparisons show that GlyphAnchor can produce complete and readable text-rich images.

\subsection{Ablation Study}
\label{sec:ablation}
\tabablation

Table~\ref{tab:ablation} presents an ablation study on the FireRed-Image-Edit-1.1 backbone.
All SFT variants are trained for the same number of steps on the same training data.
Default SFT on the same training data provides only limited gains over the original backbone.
Training with glyph patches but without positional anchoring yields substantially larger gains, confirming the value of explicit glyph conditions.
Adding positional anchoring further improves performance across all metrics, with a particularly large gain in phrase hit rate on InfoTextBench, showing that spatial grounding is crucial for making glyph patches an effective auxiliary enhancement for text-rich generation and editing.
Finally, text-aware NFT post-training yields an additional consistent boost, producing the best results on both LongTextBench and InfoTextBench.
Together, these results show that glyph priors improve text rendering and that positional anchoring is essential for making them reliable.

\section{Conclusion}
\label{sec:conclusion}
We presented \textbf{GlyphAnchor}, a method for enhancing visual text rendering in both text-to-image and image editing using position-anchored glyph priors. 
GlyphAnchor organizes glyph priors as compact glyph patch images and anchors them to target layouts through the model's native positional encoding.
This provides explicit character-shape information together with spatial cues, while keeping the conditioning mechanism lightweight and easy to adapt across diffusion-transformer backbones.
Combined with staged supervised finetuning and text-aware NFT post-training, GlyphAnchor consistently improves the corresponding backbones on long, complex, and densely arranged text and rare character scenarios. 
We also introduced \textbf{InfoTextBench} to evaluate text-rich visual text rendering. Overall, our results show that GlyphAnchor improves text fidelity while preserving image quality.

\textbf{Broader Impacts.}
While our work improves visual text generation, it may also introduce misuse risks, such as forged documents, misleading advertisements, and disinformation. Like other image generation models, GlyphAnchor may render harmful text, including obscene, hateful, or illegal content, at specified locations. Potential safeguards include content moderation, watermarking, detection tools, and audit logging, while collaboration with fact-checking organizations remains a possible future direction. We encourage responsible use and further study of safeguards for such systems.

\newpage
\acksection
This research was supported by the National Natural Science Foundation of China (NSFC 62576103). The computations were conducted using the CFFF platform at Fudan University. Part of this work was carried out during an internship at Xiaohongshu.

\bibliographystyle{plainnat}
\bibliography{main}

\clearpage
\appendix
\section{Technical appendices and supplementary material}
\subsection{Qualitative ablation on positional anchoring}
\label{app:position_anchor_qualitative}
\begin{figure}[h]
  \centering
  {\includegraphics[clip, trim=0cm 67cm 59cm 0cm, width=0.7\textwidth]{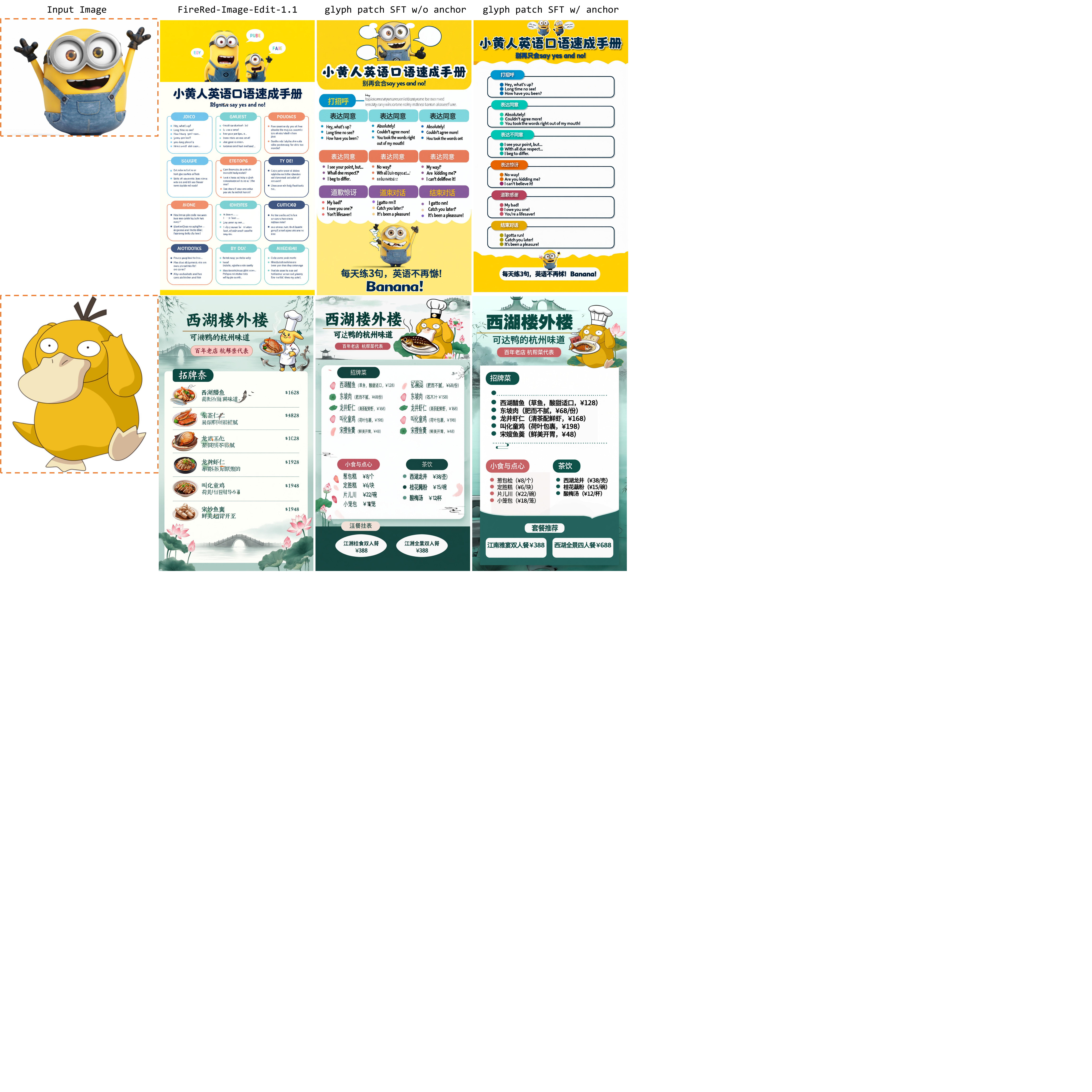}}
  \caption{\textbf{Qualitative ablation of positional anchoring.} Best viewed with zoom.}  \label{fig:qual_anchor_ablation}
\end{figure}

Fig.~\ref{fig:qual_anchor_ablation} provides a qualitative comparison of glyph patch SFT with and without positional anchoring on the FireRed-Image-Edit-1.1 backbone.
Both GlyphAnchor variants use the same staged SFT strategy, training data, rendered glyph patch images, layout inputs, and inference pipeline.
They differ only in whether glyph tokens are assigned position IDs according to their layout anchors.
Consistent with the quantitative ablation in Table~\ref{tab:ablation}, text accuracy improves progressively from the backbone to glyph patch SFT without anchoring and further to position-anchored glyph patch SFT.
Without anchoring, glyph patches are only weakly associated with the target regions, often leading to incomplete text or corrupted characters; positional anchoring instead aligns glyph priors with their target layout, enabling more accurate rendering.

\subsection{Qualitative ablation on staged SFT}
\begin{figure}[!h]
  \centering
  {\includegraphics[clip, trim=0cm 25cm 79cm 0cm, width=0.8\textwidth]{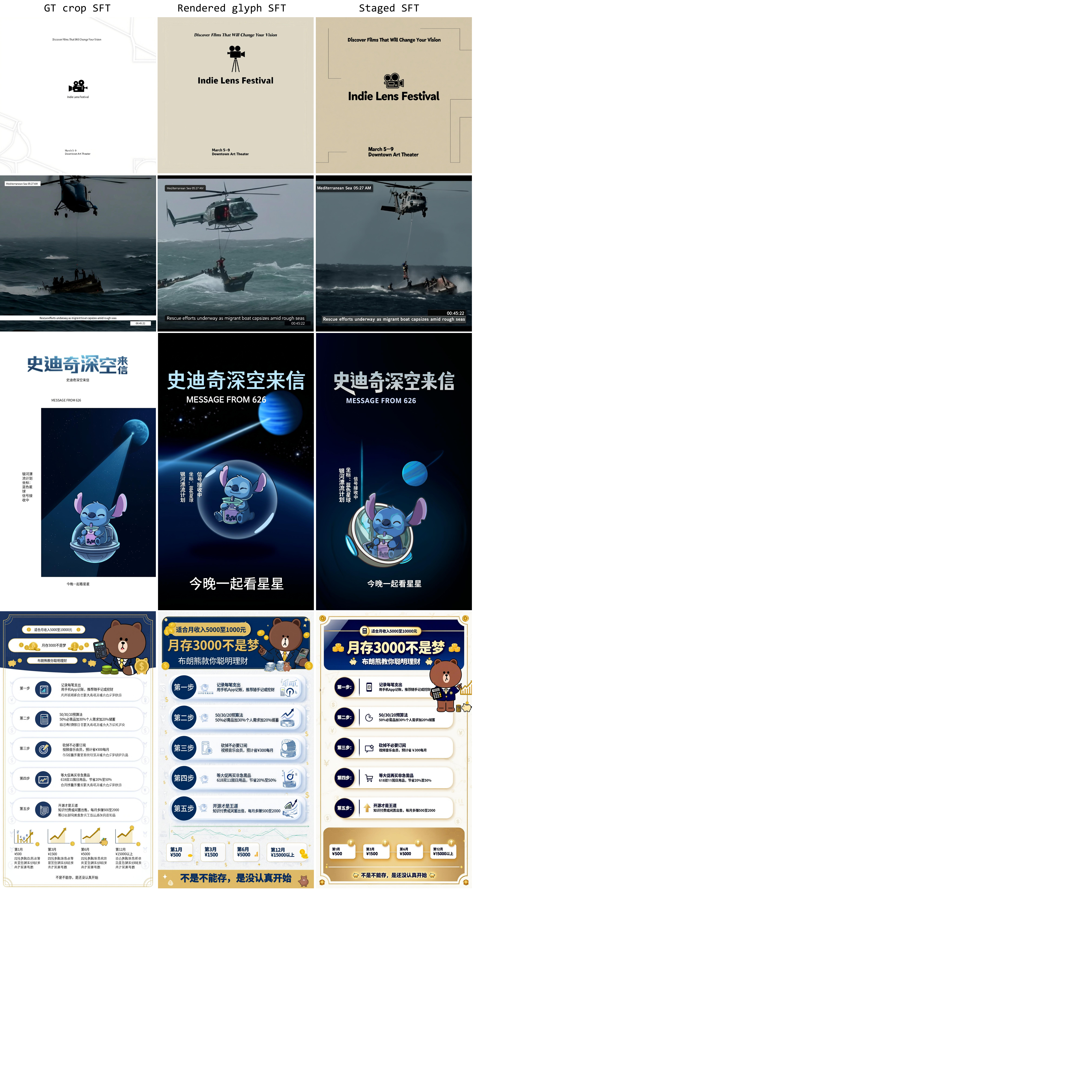}}
  \caption{\textbf{Qualitative ablation of staged SFT.} Best viewed with zoom.}
  \label{fig:qual_sft_ablation}
\end{figure}

Fig.~\ref{fig:qual_sft_ablation} qualitatively compares three SFT variants of GlyphAnchor.
All variants use FireRed-Image-Edit-1.1 as the backbone, with the same position-anchored glyph prior strategy, training data, layout inputs, and inference pipeline.
They differ only in the glyph patch images used during SFT: GT crop SFT uses cropped text regions from the ground-truth target images, rendered glyph SFT uses rendered black-on-white glyph patches, and staged SFT follows our proposed staged supervision strategy.
At inference time, all variants are conditioned on rendered glyph patches.
GT crop SFT encourages the model to transfer the appearance of the conditioning patches.
Since inference uses rendered glyph patches rather than real text crops, this supervision introduces a train--test mismatch, which manifests as white-background copy-paste artifacts.
Rendered glyph SFT alleviates this mismatch and improves visual consistency, but remains less faithful to the prompt-specified style.
In contrast, staged SFT better preserves the target image style while maintaining comparable text accuracy, yielding the most consistent overall results.

\newpage
\subsection{Robustness to Layout Perturbations}

We evaluate the robustness of GlyphAnchor to inaccurate layout estimates on InfoTextBench using FireRed-Image-Edit-1.1 as the backbone. We randomly perturb the bounding boxes by shifting their positions by up to $\pm10\%$ of their respective widths and heights and resizing them with factors in $[0.9, 1.1]$. All other generation and evaluation settings remain unchanged. As shown in Tab.~\ref{tab:layout_robustness}, GlyphAnchor maintains nearly identical performance under these perturbations, indicating robustness to moderate layout estimation errors. This observation is consistent with the train-time glyph augmentations described in Sec.~\ref{sec:staged_sft}.

\begin{table}[h]
    \centering
    \caption{Robustness to layout perturbations on InfoTextBench using FireRed-Image-Edit-1.1 as the backbone.}
    \label{tab:layout_robustness}
    \resizebox{\linewidth}{!}{
    \begin{tabular}{lcccccc}
        \toprule
        Setting & NED & Word Recall & Word Precision & Word F1 & Phrase Hit & Overall Visual Quality \\
        \midrule
        GlyphAnchor & 0.651 & 0.950 & 0.864 & 0.898 & 0.713 & 6.480 \\
        + Layout Perturbation & 0.645 & 0.955 & 0.861 & 0.898 & 0.710 & 6.456 \\
        \bottomrule
    \end{tabular}
    }
\end{table}

\subsection{Comparison of InfoTextBench and LongTextBench}
\label{app:benchmark_comparison}

Table~\ref{tab:benchmark_comparison} compares the text-length statistics of InfoTextBench and LongTextBench. InfoTextBench contains substantially longer target texts, with a mean length of 150.99 compared to 39.94 in LongTextBench. We measure text length in words for English and characters for Chinese.

\begin{table}[h]
    \centering
    \caption{Statistical comparison of LongTextBench and InfoTextBench.}
    \label{tab:benchmark_comparison}
    \begin{tabular}{lcccc}
        \toprule
        Benchmark & \# Samples & Mean Length & Median Length & Length Range \\
        \midrule
        LongTextBench & 320 & 39.94 & 33 & [10, 121] \\
        InfoTextBench & 266 & 150.99 & 135 & [18, 392] \\
        \bottomrule
    \end{tabular}
\end{table}

\subsection{Prompt Templates}
\label{app:prompt_templates}

\subsubsection{Prompt templates for NFT rewards.}
During NFT post-training in Sec.~\ref{sec:nft}, we use two MLLM prompt templates for reward computation: one for style consistency and one for visual fusion.
The style consistency prompt evaluates whether the generated text style matches the target typography and visual style, while the visual fusion prompt evaluates whether the rendered text is naturally integrated with the local background and surrounding visual context.
The Text-Layout Consistency Reward is computed separately from OCR recognition and layout matching, and therefore does not require an MLLM prompt.
The two prompt templates are provided below.

\textbf{Style consistency reward prompt template.}
\begin{tcolorbox}[
  colback=gray!10,
  colframe=gray!50,
  boxrule=0.3pt,
  arc=1mm,
  fontupper=\footnotesize,
  breakable
]
You are an expert in evaluating visual text style consistency. You are skilled at comparing the style consistency of corresponding text regions across two images. You should focus on typography, font weight, stroke shape, color, outlines, shadows, texture and material, decorative details, layout style, and the overall visual language. Your task is not to evaluate whether the text content is correct, nor to evaluate how naturally the text is fused into the image, but only to assess whether the styles of corresponding text regions are consistent across the two images.

Image A: target reference image (ground-truth image)

Image B: model-generated image

Please evaluate whether the styles of the corresponding text regions are consistent across the two images according to the following prompt. Note that different text regions within the same image may naturally have different styles, which is acceptable. You should focus on the style similarity between corresponding text regions across the two images, rather than requiring all text regions within one image to share the same style.

Prompt: [PROMPT]

If the prompt is a text removal task, such as deleting, removing, erasing, or wiping out text, this style consistency evaluation is not applicable. In that case, directly reply with 3 and do not provide any explanation.

Please consider the following aspects: typography, font weight, stroke thickness and stroke structure, handwritten vs.\ printed appearance, color, outlines, shadows, texture and material, decorative details, layout style, and overall visual tone.

Please assign an integer score from 0 to 5 according to the following criteria:

0: The styles of the corresponding text regions are completely inconsistent. The font type, color, or decorative language are clearly incorrect, and the overall result looks like a completely different design system.

1: Only a very small amount of superficial similarity is present, while the overall style is still clearly incorrect. Most key aspects (glyph form, color, outlines, brushwork, decoration) differ substantially.

2: The generated image shows some attempt to imitate the target style, but only some aspects are similar. Noticeable style drift remains, and the consistency between corresponding text regions is weak.

3: The overall style is basically similar, and the main stylistic direction is correct, but there are still differences in several important details, such as stroke characteristics, decorative treatment, color relationships, or typographic tone.

4: The styles are highly similar. Most corresponding text regions are consistent with the target image, with only minor differences that do not affect the overall style judgment.

5: The styles are almost perfectly consistent. The corresponding text regions closely match the target image in typography, stroke language, color treatment, decorative details, and overall visual tone.

Please note: evaluate only text style consistency. Do not evaluate text correctness, and do not evaluate how naturally the text blends with the background.

Please directly reply with a single integer score from 0 to 5.
\end{tcolorbox}

\textbf{Visual fusion reward prompt template.}
\begin{tcolorbox}[
  colback=gray!10,
  colframe=gray!50,
  boxrule=0.3pt,
  arc=1mm,
  fontupper=\footnotesize,
  breakable
]
You are an expert in evaluating visual text fusion in images. You are skilled at comparing the degree of harmony and integration between text and the overall image in a target reference image and a generated image. You can distinguish between real-world images and graphic design images such as posters, covers, or layout-based designs, and evaluate text fusion according to the appropriate visual standard for each case. Your task is not to evaluate whether the text content is correct, nor to evaluate whether the text style is consistent, but to assess whether the text in the generated image achieves the same level of visual fusion as in the target reference image.

Image A: target reference image (ground-truth image)

Image B: model-generated image

Please compare how well the text is integrated with the overall image in the two images, and evaluate whether the generated image reaches the level of fusion shown in the target reference image. Here, ``fusion'' refers to the degree of coordination between the text and the background, texture, lighting, perspective, edges, clarity, occlusion, surface/material attachment, and overall visual style.

If the prompt is a text removal task, such as deleting, removing, erasing, or wiping out text, this visual fusion evaluation is not applicable. In that case, directly reply with 3 and do not provide any explanation.

Prompt: [PROMPT]

Please pay special attention to the following two evaluation standards:

1. If the image is a real-world scene image (e.g., a photographed scene, a real object surface, a street view, a product image, or an environmental image), focus on whether the text looks as if it naturally exists in the scene, and whether it is consistent with realistic lighting, perspective, occlusion, material attachment, and edge details.

2. If the image is a graphic design image, poster, cover, sticker-like image, or layout-based design, focus on whether the text is visually coordinated with the design system, and whether it achieves the same level of decoration, layout quality, visual balance, and design completeness as the target image. Do not penalize it harshly based on a ``must look physically real'' standard.

Please assign an integer score from 0 to 5 according to the following criteria:

0: The text in the generated image is clearly disconnected from the overall image, and its fusion is far worse than that in the target reference image. Under either the real-world standard or the design-image standard, it looks very unnatural and awkward.

1: A few local regions may be somewhat acceptable, but the overall result is still clearly uncoordinated. There are serious problems with edge quality, spatial attachment, lighting/perspective, or design consistency.

2: The text shows partial integration with the image, but noticeable awkwardness remains. There is still a large gap compared with the target reference image.

3: The overall fusion is basically acceptable, and the major text regions look reasonably integrated. Compared with the target reference image, it reaches a moderate level, although noticeable stiffness or insufficient design coordination remains upon closer inspection.

4: The fusion quality is high. Most text regions are well coordinated with the image, and the overall effect is close to that of the target reference image, with only minor shortcomings.

5: The fusion quality reaches or almost reaches the level of the target reference image. In real-world images, the text looks very natural and believable; in design images, it looks highly coordinated and polished.

Please note: evaluate only visual fusion relative to the target reference image. Do not evaluate text correctness, and do not separately evaluate whether the text style is consistent.

Please directly reply with a single integer score from 0 to 5.
\end{tcolorbox}

\subsubsection{Prompt templates for InfoTextBench evaluation.}
For InfoTextBench in Sec.~\ref{sec:infotextbench}, we use an MLLM prompt template to evaluate overall visual quality.
The prompt asks the MLLM judge to score two aspects of each generated image: aesthetics and prompt following.
The aesthetics score reflects visual and design quality, while the prompt-following score reflects consistency with the input prompt.
The final overall visual quality score is obtained by averaging these two scores.
The prompt template is provided below.

\begin{tcolorbox}[
  colback=gray!10,
  colframe=gray!50,
  boxrule=0.3pt,
  arc=1mm,
  fontupper=\footnotesize,
  breakable
]
You are a professional digital artist and graphic designer. Your task is to evaluate an AI-generated image.

The system will provide two images:
Image A: the reference image, when applicable.
Image B: the generated image based on the reference image.

Please return your evaluation in the following format (keep the reasoning concise):

\{

"score": [aesthetic\_score, prompt\_follow\_score],

"reasoning": "..."

\}

Scoring criteria (0--10):

The first score (aesthetic, 0--10):
Evaluate the overall visual quality of the generated image, including composition and layout, color harmony, visual hierarchy, level of detail, and overall stylistic coherence.

0 = Extremely poor. The image is visually chaotic or severely flawed, with major problems such as broken composition, inconsistent or unpleasant colors, weak visual hierarchy, crude details, obvious artifacts, or a generally unfinished appearance. It does not meet even a basic visual design standard.

10 = Excellent. The image demonstrates outstanding visual quality at a highly polished professional level. The composition is well balanced, the layout is refined, the colors are harmonious and effective, the visual hierarchy is clear, the details are carefully rendered, and the overall style is highly coherent and visually impressive.

The second score (prompt following, 0--10):
Evaluate whether the generated image follows the prompt below, including layout structure, text content (whether all required text appears and is correct), style and color scheme, and the use of the reference image content.

0 = Completely fails to follow the prompt. Most key requirements are missing, incorrect, or ignored. The layout, text, style, or major visual elements do not match the prompt description in any meaningful way.

10 = Perfectly follows the prompt. All major and minor requirements are faithfully satisfied, including the intended layout, required text content, stylistic description, color design, and visual elements. The generated image closely matches the prompt in both content and presentation.

Generation prompt:
<PROMPT>

Required text list (provided to help judge whether the text is rendered correctly):
<TEXTS>
\end{tcolorbox}

\subsection{English Translation of the Prompt in Fig.~\ref{fig:glyph_prior_motivation}}
\label{app:prompt_translation}
For readability, all Chinese text, including the target text to be rendered, has been translated into English. The original prompt was entirely in Chinese.
\begin{tcolorbox}[
    colback=gray!10,
    colframe=gray!50,
    boxrule=0.3pt,
    arc=1mm,
    fontupper=\footnotesize,
    breakable
]
Use the provided Chiikawa character image to generate a healing-themed book recommendation poster. All text should be in Simplified Chinese. The canvas should be a long vertical poster. The overall style should be warm, artistic, and hand-drawn, with an off-white background as the main color, accented by soft tones such as milk-tea brown, mint green, and warm orange to create a fresh, gentle, bookish atmosphere. ...
At the very top center, place a small milk-tea-brown capsule-shaped label containing the text ``2024 Good Book Recommendations'' in a small, rounded dark-brown font, with a small book icon on the left. Below the label is the main title ``Chiikawa's Healing Reading List,'' set in a large, bold, rounded handwritten-style font in dark brown and centered horizontally. Below the main title is the subtitle ``10 Books That Make the Heart Softer,'' set in a medium-sized rounded font in warm brownish-gray and centered horizontally, with a small, slightly open book illustration on each side in mint green. The Chiikawa character should be placed centrally below the title card, sitting on top of a colorful stack of books while holding an open book with both hands. ...
Arrange 10 book entries vertically, separated by thin light-brown dashed lines. Each entry should be a light beige card with rounded corners. On the left side of each card, place a narrow vertical decorative color bar, alternating among mint green, pale yellow, pink-orange, and sky blue, with a small dot of the same color at the top. On the left side of each book card, combine the index number and book title into a single line in a medium-sized, bold, dark-brown font. Below it, place a quote in a small, rounded, medium-gray font, with a small light-brown quotation mark decoration on the left. The first book card should display ``No.1 The Little Prince,'' with the quote ``Love means responsibility, not possession.'' The second book card should display ``No.2 To Live,'' with the quote ``People live for the sake of living itself.'' The third book card should display ``No.3 The Miracles of the Namiya General Store,'' with the quote ``Every trouble has an answer; you just do not know it yet.'' The fourth book card should display ``No.4 Norwegian Wood,'' with the quote ``Death is a part of life.'' The fifth book card should display ``No.5 The Storied Life of A.J. Fikry,'' with the quote ``No one is an island.'' The sixth book card should display ``No.6 Ferryman,'' with the quote ``You are the most beautiful miracle in my wasteland.'' The seventh book card should display ``No.7 The Moon and Sixpence,'' with the quote ``Look up at the moon, but also pick up the sixpence.'' The eighth book card should display ``No.8 The Courage to Be Disliked,'' with the quote ``All troubles come from interpersonal relationships.'' The ninth book card should display ``No.9 One Hundred Years of Solitude,'' with the quote ``There are not many things in life that are truly important.'' The tenth book card should display ``No.10 Hello, Old Times,'' with the quote ``May you encounter beautiful moments.'' To the right of each book title, place a small rounded decorative color block corresponding to the color palette of that book's cover.
The concluding section at the bottom should occupy approximately one-eighth of the canvas height. ... Center a line of text in this area reading ``May every book be a bridge that gently reconciles you with the world,'' in a medium-sized, bold, rounded font in creamy white, with a small heart icon on each side, also in creamy white.
\end{tcolorbox}


\end{document}